\documentclass[10pt,twocolumn,letterpaper]{article}

\usepackage{cvpr}              %

\usepackage[accsupp]{axessibility} %
\usepackage{graphicx}
\usepackage{amsmath}
\usepackage{amssymb}
\usepackage{booktabs}
\usepackage{multicol}
\usepackage{amsthm}
\newtheorem{prop}{Proposition}
\usepackage{multirow}
\usepackage[acronym]{glossaries}
\newacronym{elbo}{ELBO}{Evidence Lower Bound}
\newacronym{mc}{MC}{Monte Carlo}
\newacronym{hmc}{HMC}{Hamiltonian Monte Carlo}
\newacronym{bnn}{BNN}{Bayesian Neural Network}
\newacronym{cnn}{CNN}{Convolutional Neural Network}
\newacronym{sgld}{SGLD}{Stochastic Gradient Langevin Dynamics}
\newacronym{fcn}{FCN}{Fully Convolutional Networks}
\newacronym{duq}{DUQ}{Deterministic Uncertainty Quantification}
\newacronym{ep}{EP}{Expectation Propegation}
\newacronym{gp}{GP}{Gaussian Process}

\newcommand{\x}[0]{\mathbf{x}}
\newcommand{\y}[0]{\mathbf{y}}
\newcommand{\w}[0]{\boldsymbol{\omega}}
\newcommand{\X}[0]{\mathbf{X}}

\newcommand{\Y}[0]{\mathbf{Y}}

\newcommand{\T}[0]{\mathbf{t}}
\newcommand{\N}[0]{\mathcal{N}}
\newcommand{\E}[0]{\mathbb{E}}

\newcommand{\sm}[1]{\text{Softmax} (#1)}

\usepackage{tabularx}
\usepackage{booktabs}
\newcolumntype{L}[1]{>{\raggedright\let\newline\\\arraybackslash\hspace{0pt}}m{#1}}
\newcolumntype{C}[1]{>{\centering\let\newline\\\arraybackslash\hspace{0pt}}m{#1}}
\newcolumntype{R}[1]{>{\raggedleft\let\newline\\\arraybackslash\hspace{0pt}}m{#1}}

\usepackage[pagebackref,breaklinks,colorlinks]{hyperref}
\usepackage{xcolor}
\usepackage[capitalize]{cleveref}
\crefname{section}{Sec.}{Secs.}
\Crefname{section}{Section}{Sections}
\Crefname{table}{Table}{Tables}
\crefname{table}{Tab.}{Tabs.}

\def\cvprPaperID{5} %
\def\confName{CVPR}
\def\confYear{2023}

\begin{document}

\title{Uncertainty in Real-Time Semantic Segmentation on Embedded Systems}

\author{Ethan Goan, Clinton Fookes \\
Queensland University of Technology\\
Queensland, Australia\\
{\tt\small ej.goan@qut.edu.au}
}
\maketitle
\begin{abstract}
Application for semantic segmentation models in areas such as autonomous
vehicles and human computer interaction require real-time predictive
capabilities. The challenges of addressing real-time application is amplified by
the need to operate on resource constrained hardware. Whilst development of
real-time methods for these platforms has increased, these models are unable to
sufficiently reason about uncertainty present when applied on embedded real-time systems. This paper addresses this by
combining deep feature extraction from pre-trained models with Bayesian
regression and moment propagation for uncertainty aware predictions. We
demonstrate how the proposed method can yield meaningful epistemic uncertainty on embedded
hardware in real-time whilst maintaining predictive performance.
\end{abstract}
\section{Introduction}
\label{sec:intro}
Development and capabilities of semantic segmentation models has increased
dramatically, with models based on deep learning applied to domains such as
autonomous vehicles
\cite{brostow2009semantic,caesar2018coco,cordts2016cityscapes}, robotics \cite{milioto2018real} and human
computer interaction \cite{zhou2017scene,caesar2018coco}. Practical application in these areas requires systems to provide real-time performance, and in safety critical domains meaningful uncertainty information is essential. Providing this uncertainty information becomes increasingly challenging for real-time operation, as obtaining this uncertainty information can considerably increase compute demands. A natural way to represent epistemic uncertainty is through a Bayesian framework, where uncertainty in the model is propagated to predictions.
\par
Complete Bayesian inference is intractable for deep learning models, meaning
expensive Monte Carlo integration is often used for approximate inference
\cite{mukhoti2018evaluating,dechesne2021bayesian,kampffmeyer2016semantic}. These sampling based
techniques considerably increase the time required for prediction, making them
unsuitable for real-time operation. Other research have proposed utilising uncertainty
information during training \cite{sivaprasad2021uncertainty,zheng2021rectifying}, though
are unable to sufficiently reason about epistemic uncertainty present during predictions. Given that the primary practical application of many semantic segmentation
models is performed on edge computing devices, it is crucial that we are able to
 deliver this uncertainty information on these platforms in real time.
\par
\begin{figure}[t]
\begin{minipage}[b]{.48\linewidth}
  \centering
\centerline{\includegraphics[width=4cm]{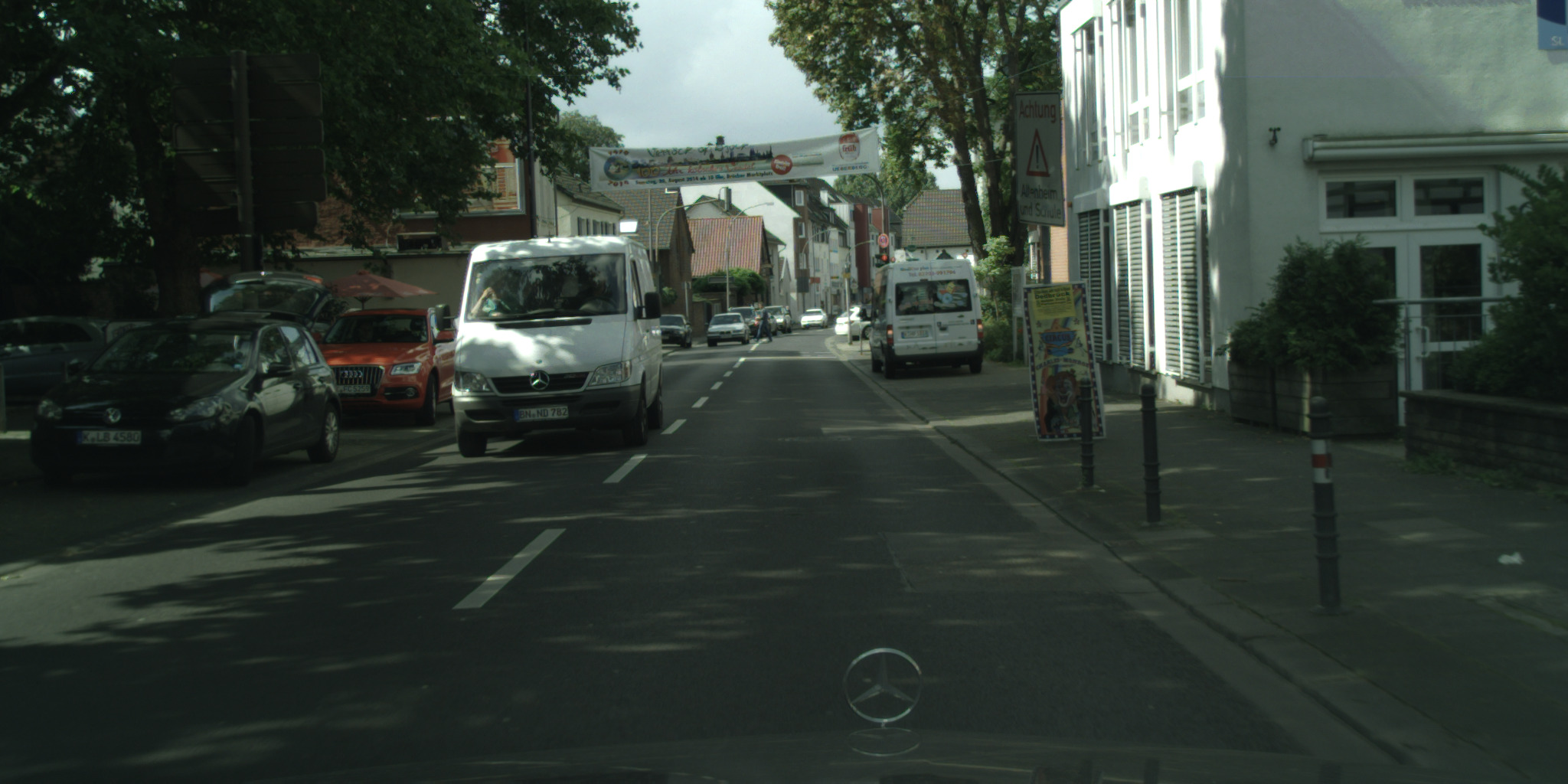}}
  \centerline{input image}\medskip
\end{minipage}
\hfill
\begin{minipage}[b]{0.48\linewidth}
  \centering
\centerline{\includegraphics[width=4cm]{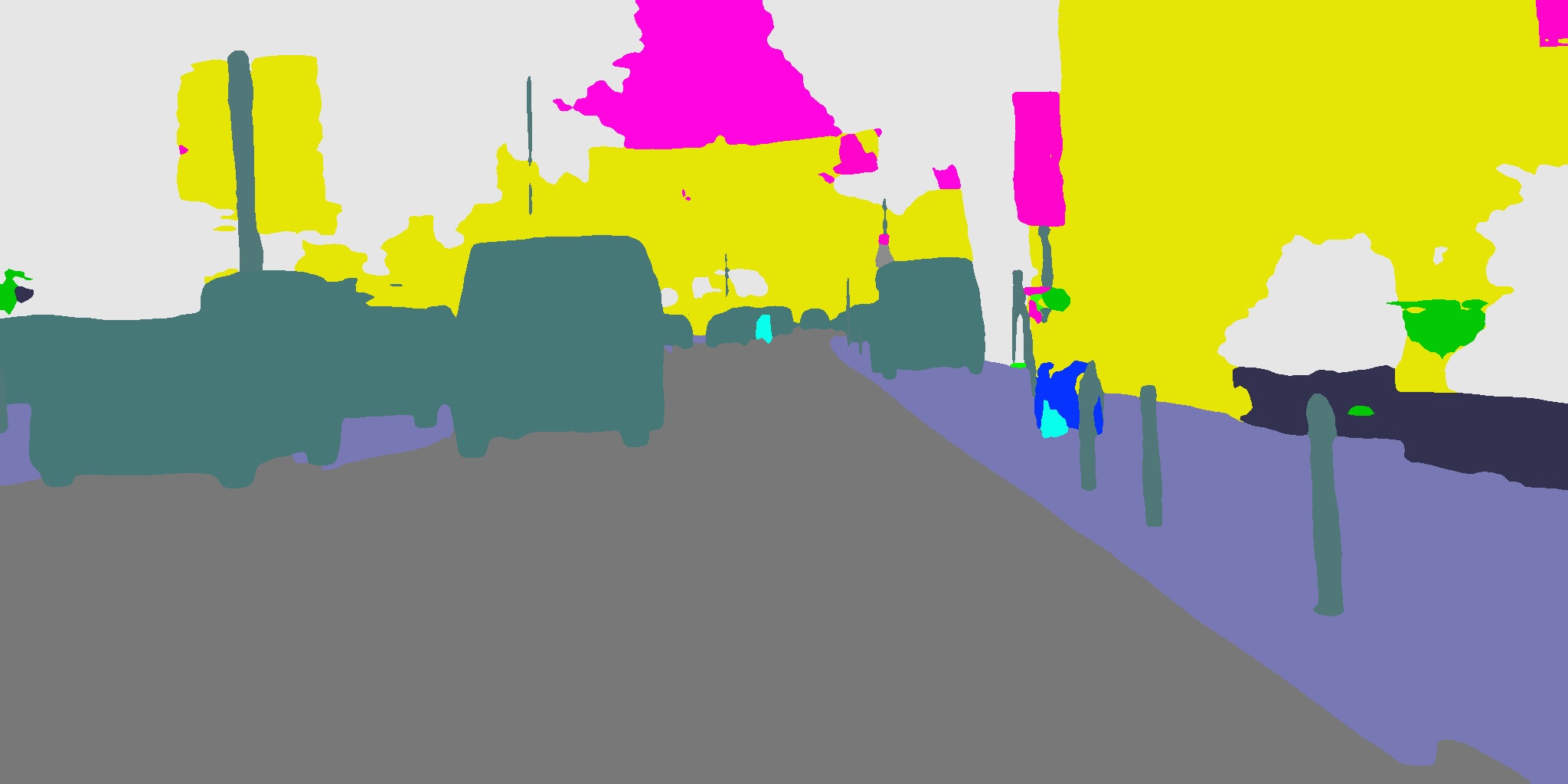}}
  \centerline{segmentation}\medskip
\end{minipage}
\begin{minipage}[b]{.48\linewidth}
  \centering
\centerline{\includegraphics[width=4cm]{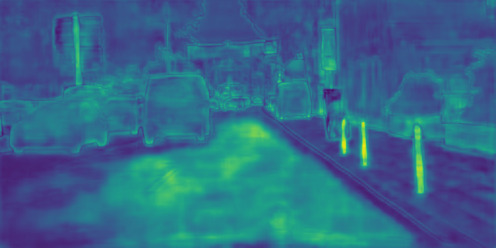}}
  \centerline{epistemic uncertainty}\medskip
\end{minipage}
\hfill
\begin{minipage}[b]{0.48\linewidth}
  \centering
\centerline{\includegraphics[width=4cm]{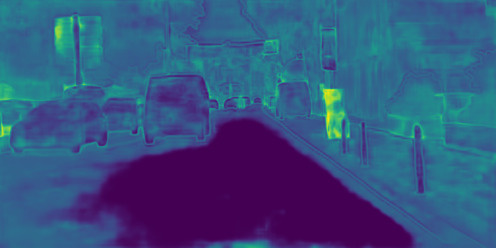}}
  \centerline{aleatoric uncertainty}\medskip
\end{minipage}
\caption{Example of semantic segmentation results obtainable from proposed real-time semantic segmentation method and the forms of uncertainty it permits in real-time applications.}
\label{fig:res}
\end{figure}
This work aims to address this through the use of an analytic probabilistic module inspired by the work of the Gaussian Process treatments of deep kernel learning and Bayesian optimization \cite{wilson2016deep, snoek2015scalable}, where the task of feature extraction is computed deterministically with fixed model parameters, which are then used as inputs to a probabilistic classification module. Complexity for Gaussian processes is cubic in the number of training samples, thus making them infeasible for the intended applications. Instead we opt for parameterised probabilistic regression module combined with a moment propagating non-linearity for classification. This reduces the inference overhead for high-dimensional data whilst providing analytic results when using conjugate models, avoiding the need for expensive Monte Carlo approximations. We demonstrate how this approach can be applied to existing real-time semantic segmentation models on embedded hardware whilst providing meaningful uncertainty measures with minimal compute overhead. An example of the predictions and uncertainty measures for the presented models is shown in~\Cref{fig:res}.
The contributions of this paper are as follows,
\begin{itemize}
  \item Propose a light-weight method for uncertainty in semantic segmentation by combining Bayesian methods within moment propagation,
  \item  Develop meaningful real-time aleatoric and epistemic uncertainty metrics and investigation into how they can inform end users and decision protocols,
  \item Demonstrate how these methods can be easily adapted to pre-trained models, enabling them to provide real time uncertainty quantification on embedded devices.
\end{itemize}

\section{Related Work}
\subsection{Deep Learning Models for Real-Time Semantic Segmentation}
 Whilst state-of-the-art methods in terms of predictive performance utilise modern deep-learning architectures such as transformers \cite{chen2022vision,yuan2019segmentation} or large \Gls{fcn} models \cite{wang2022internimage,fu2020scene}, real-time methods are restricted to a class of computationally efficient methods that allow them to be deployed in practice.
Real-time semantic segmentation models utilising deep learning architectures
traditionally build upon \cite{long2015fully} by using \Gls{fcn} architectures consisting of both an encoder and decoder. Foundational real-time methods include ENet \cite{paszke2016enet} which provides a lightweight \Gls{fcn} utilising skip-connections and bottleneck layers similar to ResNets \cite{he2016deep}, and max-unpooling in the decoder stage. This lightweight model has shown to provide fast predictive performance on consumer hardware at the expense of accuracy. Modern real-time models extend upon this work to design more complex encoders and decoders \cite{wang2021swiftnet} or through the inclusion of additional auxilliary loss functions that aim to capture semantic information over a larger receptive field\cite{yu-2018-bisen,yu-2020-bisen-v2,xu2022pidnet}.
\subsection{Uncertainty in Semantic Segmentation}
Description of epistemic uncertainty in semantic segmentation methods typically rely
on sampling based procedures such as Monte-Carlo Dropout
\cite{kampffmeyer2016semantic,kendall2017bayes,kendall2017uncertainties,mukhoti2018evaluating}. To
obtain uncertainty information with these methods, a single sample must be
passed through the model multiple times whilst sampling model parameters from
the approximate posterior. These factors restrict their application to offline
methods where considerable compute resources are available. \Gls{duq} methods
avoid the need for multiple forward passes during test time, though requires the
inclusion of an additional weight matrix for each class present in the dataset
and a Gaussian Discriminant Model to be placed afterwards
\cite{van2020uncertainty,mukhoti2021deep}. \cite{huang2018efficient}
provide an uncertainty mechanism building upon Monte-Carlo Dropout that
incorporates averages of an optical flow model to avoid multiple forward passes
through frames, though is only applicable to video capture scenarios and relies
on aggregating samples over multiple time-steps. This aggregation increases
memory footprint for predictions, as prior samples from the optical flow model
need to be stored, and computational load is increased due to the need to
calculate uncertainty metrics over samples within the flow model. None of the
methods reviewed here have
been shown to offer real-time predictive performance on traditional hardware,
and to the best of our knowledge no prior work has investigated real-time epistemic
uncertainty estimation on resource constrained embedded devices.
\subsection{Gaussian Process Methods and Expectation Propagation}
To address the limitations of existing uncertainty quantification for semantic
segmentation, we will build upon the work of Deep Gaussian Process models
and \Gls{ep}.
Deep Gaussian Process models such as \cite{wilson2016deep,snoek2015scalable} separate the tasks of feature extraction and
classification, where a \Gls{gp} model is utilised for probabilistic
classification. Whilst \Glspl{gp} reduce the need for multiple forward passes,
their application requires a large number of parameters to approximate the
corresponding covariance function. \cite{qu2021bayesian} use a similar concept
for separating feature extraction and probabilistic classification for monocular
depth estimation. \Gls{ep} methods \cite{frey1999variational,gast2018light} similarly avoid the need for
multiple forward passes, where the distribution of a random variable is
approximated analytically through the different transformations. We build upon
the concepts of probabilistic classification using deep features and \Gls{ep} to
enable uncertainty quantification for semantic segmentation in real--time on
embedded hardware, which we describe in the following section.

\section{Real-time Uncertainty Quantification}
\begin{figure}%
\centering
\includegraphics[width=1\columnwidth]{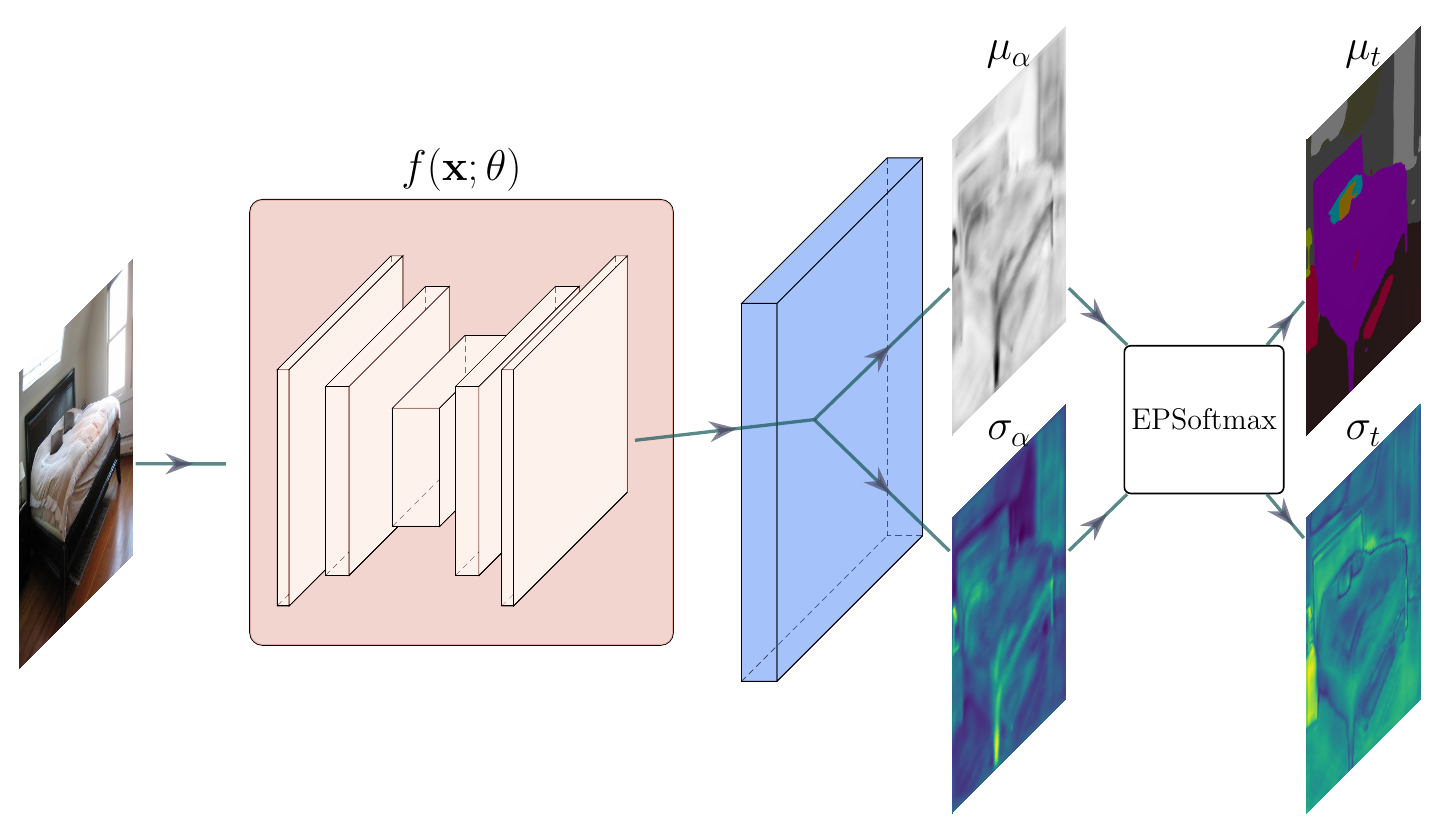}%
\caption{Summary of the proposed model. The pink box represent the deterministic feature extractor $f(\x; \theta)$ that can be any pretrained semantic segmentation network. The blue convolutional layer represents our probabilistic layer, which allows for analytic predictive inference, and is used to generate a mean and variance representation of the predictive logits \(\alpha\). These are then passed  through the EPSoftmax layer to represent final predictive categories $\mathbf{t}$ and epistemic uncertainty information.}
\label{fig:network}
\end{figure}%

We pose the quantification of epistemic uncertainty within the Bayesian
framework, where we propagate uncertainty in the model parameters to
predictions. For this we require a posterior of our model latent
variables $\w$ such that,

\begin{equation}
  \label{eq:bayes} p(\w | \X, \Y) \propto p(\w) p(\X, \Y | \w)
\end{equation}
where $\X$ is the set of our training inputs and $\Y$ is a matrix of our predictive labels.
We can then form a predictive distribution for new test data $(\hat{\x}, \hat{\y})$ as,
\begin{equation}
  \label{eq:bayes_pred} p(\hat{y}| \hat{\x}, \X, \Y) = \int p(\w | \X, \Y) p(\hat{\x}, \hat{\y} | \w) d\w.
\end{equation}
For the case of a neural network representing our likelihood, the
integrals required for Eqns. \eqref{eq:bayes} and \eqref{eq:bayes_pred} are intractable when
treating all network parameters as random variables. For the case
of \eqref{eq:bayes}, the posterior may be approximated by a tractable
form through means such as variational inference, though the predictive posterior in \eqref{eq:bayes_pred} for deep
neural networks remains intractable and requires expensive Monte-Carlo integration to be used.
\par
The goal of this work is to find a simpler probabilistic representation of model predictions
for semantic segmentation that provides efficient analytic computation, thus avoiding the need for expensive sampling approximations. To achieve
this, we propose simplifying the probabilistic component of our model to that of
linear model with respect to latent variables, where the basis function applied to the input is a neural
network $f(\cdot; \theta)$ parameterised by $\theta$ which includes all but the last layer of the neural
network. These parameters are treated as fixed and known values. We then replace the final linear layer within a given network with a probabilistic layer parameterised by our latent variables $\w$ for which we will perform inference. This can
be seen as separating the model into a feature extraction stage with fixed
parameters $\theta$, and the classification stage with parameters treated as random
variables \(\w\). The latent parameters in $\w$ will represent the weights and bias of a final convolutional layer applied to the model.~\Cref{fig:network} illustrates the proposed model, and can be summarised as,
\begin{align}
  \label{eq:pred}
  \Phi(\x) &= f(\x; \theta)
  \\
  \alpha &= \Phi(\x) \w
  \\
  \T &= \sm{\alpha},
\end{align}
where $\Phi(\x)$ represents the design matrix for input $\x$ generated from a
neural network with fixed parameters $\theta$,  $\alpha$ and $\T$ represents the
corresponding logits and final
predictive categorical probability respectively, and $\w \sim p(\w| \X, \Y, \theta)$ is
the latent variables we wish to perform inference for. We represent the multiplication between the design matrix and the probabilistic parameters as an inner product, but note that this inner product can be computed with an equivalent convolution operation. Since inference is only required for the final latent variables, we are able to leverage pretrained networks $f(\x; \theta)$ for the generation of design matrices.
\par
For the work presented within, we frame
inference in the logit space as opposed to the final categorical distribution.
This allows us to frame the inference problem as a simple linear regression
model, such that our posterior of latent variables $\w$ with $N$ data points is,
\begin{align}
  \label{eq:prob}
 p(\w | \X, \Y , \theta ) \propto \Big[\prod_{i=0}^{N-1}\N\Big(y|f(\x_{i};\theta) \w, \sigma^{2}\Big)\Big] p(\w).
\end{align}
With a conjugate Gaussian prior placed on latent variables $\w$, our posterior
will also be Gaussian \cite{murphey2012machine},
\begin{equation}
  \label{eq:post}
  p(\w | \X, \Y , \theta ) = \N(\mu_{\pi}, \Sigma_{\pi}).
\end{equation}

The advantage of representing our model in this way is that it allows us to
represent predictive probabilities over the logits for a single test image $\hat{\X}$ analytically as,
\begin{align}
  \label{eq:2}
  p(\hat{\alpha}|\hat{\x}, \X, \Y, \theta)  &=
\int  \N \Big(\Phi(\hat{\x}) \w , \sigma^{2}\Big)  p(\w | \X, \Y , \theta ) d\w
\\
& = \N(\hat{\alpha}| \Phi(\hat{\x})  \mu_{\pi}, \sigma^{2}_{N}(\hat{\x}))
  \\
\label{eq:pred_cov}
 \sigma^{2}_{N}(\hat{\x})& = \sigma^{2} + \Phi(\hat{\x})^{T}\Sigma_{\pi}\Phi(\hat{\x})
\end{align}

\subsection{Inference of Latent Variables}
\label{sec:inference}
A challenge with this modelling scheme is that since we are performing inference
in the logit space, we cannot directly define a likelihood based on the data, as
the labels for our data is represented as final categorical probabilites and
the softmax function applied to logits to obtain this predictive probability is
not bijective. To address this, we instead propose to approximate samples from
the  conditional posterior of our latent variables $\w$ using the diagonal SWAG
method \cite{maddox2019simple}.
\par
Given a conjugate prior, we know that the posterior of our latent
variables $\w$ will be a Gaussian, meaning it can be fully described using
only the first two moments. To summarise these moments, we apply the diagonal SWAG method
to approximately sample from our model parameters using SGD iterates to compute $\w_{\text{SWA}} =\frac{1}{T} \sum_{i=1}^{T} \w_{i}$. From this, we can compute an online empirical variance estimate for our probabilistic parameters as $\Sigma_{\text{diag}} = \text{diag}(\bar{\w}^{2} - \w_{\text{SWG}}^{2} )$. In this work, we differ from the original SWAG method, where instead of computing the empirical mean from the SWAG iterates, we use the pretrained weights for the last convolutional layer in our network $\bar{\w} = \w_{\text{pretrained}}$. This reduces the need to record many parameter values during training, and encourages mean predictions from the probabilistic model to be similar to the point estimate network. With this empirical variance parameters,  we can represent
our approximate posterior over our $N$ probabilistic parameters as a factorised Gaussian distribution,
\begin{equation}
  \label{eq:5}
p(\w | \X, \Y , \theta ) \approx \prod_{i=1}^{N}  \N(\w_{i} | \bar{\w}, \Sigma_{\text{diag}} ).
\end{equation}
This allows us to perform approximate inference in the logit space by utilising the
categorical labels available for our data sets. Furthermore, the use of a
factorised approximation accelerates computation of predictive posterior, as the
covariance matrix $\Sigma_{\pi}$ in~\eqref{eq:pred_cov} becomes a diagonal matrix, such that,
\begin{align}
  \label{eq:pred_approx}
 \sigma^{2}_{N}(\hat{\x})& \approx \sigma^{2} + \Phi(\hat{\x})^{T}\Sigma_{\text{diag}}\Phi(\hat{\x}).
\end{align}

Whilst eliminating covariance within our posterior approximation, the use of
a diagonal covariance matrix considerably reduces the number of model
parameters, allowing for easier adaption to low
memory applications and accelerate prediction on resource constrained devices.
\subsection{Predictive Distributions and Measures of Uncertainty}
Whilst providing an analytic solution to the predictive probabilities of our
logits, we ultimately wish to represent a predictive distribution for final
categorical probabilities. An exact analytical solution when using the softmax
activation is not known. We address this by building on the methods of
moment propagation used in the works of \Gls{ep}, where instead
of computing exact distributions, we approximate the moments of random variables
after applying certain functions, and use these moments to approximate the
transformed random variables as Gaussians. Similar to \cite{gast2018light}, we compute
intermediate moments of the softmax using properties of
the Log-Normal distribution.
\par
For a Gaussian random
variable
$\gamma \sim \N(\mu_{\gamma}, \sigma_{\gamma}^{2})$,
$\beta = \exp(\gamma) \sim \text{Lognormal} (\mu_{\beta},\sigma_{\beta}^{2})$, where the mean and
variance of $\beta$ is  $\mu_{\beta} = \exp\big(\mu_{\gamma} + \sigma_{\gamma}^{2} / 2\big)$
and $\sigma^{2}_{\beta} = \big(\exp{\sigma_{\gamma}^{2}} -1\big)\exp(2\mu_{\gamma} + \sigma_{\gamma}^{2})$ respectively. We can use
these two moments to approximate the transformed random variables as a
Gaussian $\beta \dot{\sim}\N(\mu_{\beta}, \sigma^{2}_{\beta})$. From this we can approximate the
distribution of the output for each class as,
\begin{equation}
  \label{eq:softmax_prob_start}
  t_{j} = \frac{\exp(\alpha_{j})}{\sum_{i=0}^{N-1}\exp(\alpha_{i})} =
 \frac{y_{j}}{\sum_{i=0}^{N-1}y_{i}}
\end{equation}
where $\alpha_{j}$ is a single logit from our probabilistic layer and $y_{i} \sim \N(\mu_{y, i}, \sigma^{2}_{y, i})$ using the properties of the Log-Normal distribution. To find the mean and variance for
our output $t_{i}$ using moment propagation, we first find the mean and variance of the denominator.
Assuming independence in our outputs, the denominator distribution in~\Cref{eq:softmax_prob_start} can be represented as,
\begin{equation}
  \label{eq:denominator_dist}
\sum_{i=0}^{N-1}y_{i} \sim \N(\sum \mu_{y,i}, \sum \sigma^{2}_{y,i}) = \N(\mu_{d}, \sigma^{2}_{d}).
\end{equation}
\par
The application of moment propagation used in \cite{gast2018light} does not aim to approximate the moments from the ratio distribution encountered within a probabilistic treatment of the softmax, and instead reduce the Gaussian random variable to a Dirichlet random variable, meaning additional computation is required to obtain class-conditional uncertainty measures. We rectify this by following ratio distribution approximation of \cite{diaz2013existence}, where a Gaussian approximation is derived from a Taylor expansion. With this we can approximate the mean and variance of each softmax output indexed by $j$ as\footnote{The variance from \cite{diaz2013existence} is $\mu_{j}^{2} /\mu_{d}^{2}(\sigma^{2}_{j} / \mu_{j}^{2} +  \sigma^{2}_{d} / \mu_{d}^{2})$, though this is more prone to overflow in half-precision floating point.},
\begin{equation}
  \label{eq:pred_prob}
t_{j} \dot{\sim} \N\Big(\frac{\mu_{j}}{\mu_{d}},  \frac{\mu_{j}}{\mu_{d}}\sqrt{\sigma^{2}_{j} + \sigma^{2}_{d}}\Big)
\end{equation}
where $\mu_{j}$ and $\sigma_{j}$ are found from the Log-Normal properties.
We label the use of this moment propagation as the Expectation Propagation  Softmax (EPSoftmax). 
\par
Whilst having the ability to represent classification uncertainty conditional on our deep feature extraction, we
also wish to form a suitable categorical distribution over outputs for
classification using the conventional $\text{argmax}$ operator. We show how we can obtain categorical probabilities for classification using the described approach.
\begin{prop}
  \label{prop:categorical}
  With the probabilistic output in \cref{eq:pred_prob}, we can create a
  valid $k-$dimensional categorical distribution such that,
\begin{equation}
  \label{eq:categorical}
 C \sim \text{Cat}(k,\E[\mathbf{t}]).
\end{equation}
  \end{prop}
\begin{proof}
For $\text{Cat}(k,\E[\mathbf{t}])$ to be a valid categorical distribution, we
require that the parameters generated by $\E[\mathbf{t}]$ satisfy the
condition that $\E[\mathbf{t}]_{j} \leq 1$ for $0 \leq j < k $  and  $\sum_{i=0}^{k-1} \E[\mathbf{t}]_{i} = 1$. We first observe that
expectation for the $j^{th}$ component is,
\begin{equation}
  \label{eq:4}
  \E[\mathbf{t}]_{j} =  \frac{\mu_{j}}{\mu_{d}} =
  \frac{\exp\big(\mu_{j} + \sigma^{2}_{j} / 2\big)}{\sum_{i=0}^{k-1} \exp\big(\mu_{i} + \sigma^{2}_{i} / 2\big)} =   \frac{\exp\big(\mu_{j} + \sigma^{2}_{j} / 2\big)}{\mu_{d}},
\end{equation}
where properties of the Lognormal distribution have been used and $\mu_{d}$ is that of~\Cref{eq:denominator_dist}.
Similar to the traditional softmax function, the denominator is shared amongst
all components, and the exponential function ensures all components are strictly
positive. Following the same arguments of the softmax, this implies
that $\exp\big(\mu_{j} + \sigma^{2}_{j} / 2\big)$ for valid index $j$, and
that $\sum_{i}^{k-1}\E[\mathbf{t}]_{i} = 1$ as required.
\end{proof}
With this categorical representation, we can easily perform final classification
similar to point estimate networks, whilst also delivering important uncertainty
information in these predictions. In the next section, we discuss the
uncertainty metrics available.
\subsection{Computing Predictive Uncertainty}
\label{rec:comp_pred_uncertainty}
With our representation of predictive probabilities, we want to be able to
reason about the different types of uncertainty present. It is common to refer
to the epistemic and aleatoric uncertainty \cite{kendall2017uncertainties},
where epistemic uncertainty is a representation of reducible uncertainty
identified by the model, and aleatoric uncertainty is irreducible uncertainty
caused by the data present. With the probabilistic model presented within, we
are able to reason about both forms of uncertainty.
\par
We propose to capture epistemic uncertainty using the variance information
available in our model predictions. An advantage of our approach is that we can
measure this uncertainty in both the logit and prediction space. Given we are
modelling these spaces as Gaussians, we can
succinctly summarise this uncertainty using the entropy of these distributions.
The entropy for a D-dimensional multivariate Gaussian is,
\begin{equation}
  \label{eq:entropy_gaussian}
  \mathbb{H}[x] =  \frac{1}{2}\log |\Sigma| + \frac{D}{2}(1 + \log 2\pi).
\end{equation}
We see from \cref{eq:entropy_gaussian} that the entropy depends only
on the covariance information in the distribution. This is advantageous as it
provides a succinct and natural form to summarise the uncertainty induced by our
latent variables in our predictions. Furthermore, given that we are representing
our distributions over our logits and final predictions as independent Gaussian
variables, the determinant operator can be reduced to a product of the variance
components for each component in the relevant distribution.
\par
Entropy provides a succinct summary for our uncertainty over all categories of interest,
though there are instances where we may wish to view uncertainty information
pertaining to a single class of interest. For example, for an autonomous vehicle
scenario we may be interested in observing the uncertainty for safety critical
classes such as other vehicles or pedestrians. We can summarise the class
conditional uncertainty utilising the corresponding variance in our marginal predictive probability
in~\cref{eq:pred_prob} for classes of interest.
\par
It has been shown that entropy of the final categorical probability obtained
from classification models can serve as an upper-bound for aleatoric
uncertainty \cite{smith2018understanding}. We build upon this insight in this
work to allow us to approximate aleatoric uncertainty $\mathcal{U}_{A}$ using the entropy $\mathbb{H}$ of the categorical distribution represented
in~\eqref{eq:categorical} such that,
\begin{equation}
  \label{eq:aleatoric_uncertainty}
 \mathcal{U}_{A} \approx  \mathbb{H}[C]
\end{equation}
Using the method
described here, we are able to reason about both aleatoric uncertainty and
epistemic uncertainty present within our model. We demonstrate in the following
section how we can use the described approach for real-time semantic segmentation.

\section{Experiments}
\label{sec:experiments}
\subsection{Experimental Setup}
With our probabilistic model, inference scheme and uncertainty measures defined,
we now demonstrate how they can all be combined to deliver uncertainty in
real-time compute constrained devices. The NVIDIA Jetson AGX Xavier embedded
device is used as the computing platform. Our aim here is to demonstrate
how the proposed uncertainty mechanism can be applied to existing pretrained
semenatic segmentation models to endow them with the capability to compute
aleatoric uncertainty measures. Code for experiments is available at \url{https://github.com/egstatsml/eu-seg}.
\begin{figure*}%
\centering
\begin{subfigure}{.4\columnwidth}
\includegraphics[height=2.5cm,width=\columnwidth]{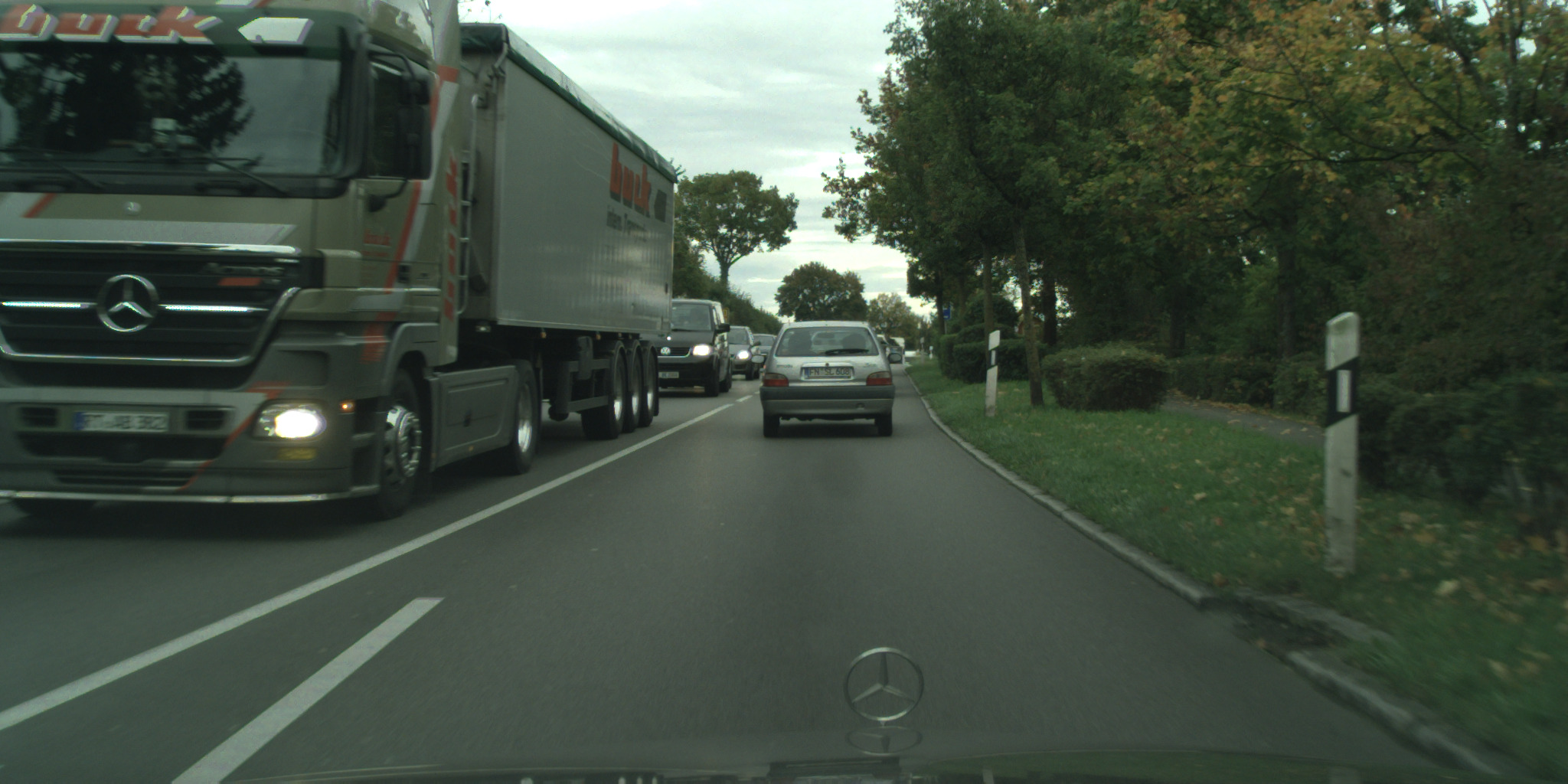}%
\end{subfigure}\hfill%
\begin{subfigure}{.4\columnwidth}
\includegraphics[height=2.5cm,width=\columnwidth]{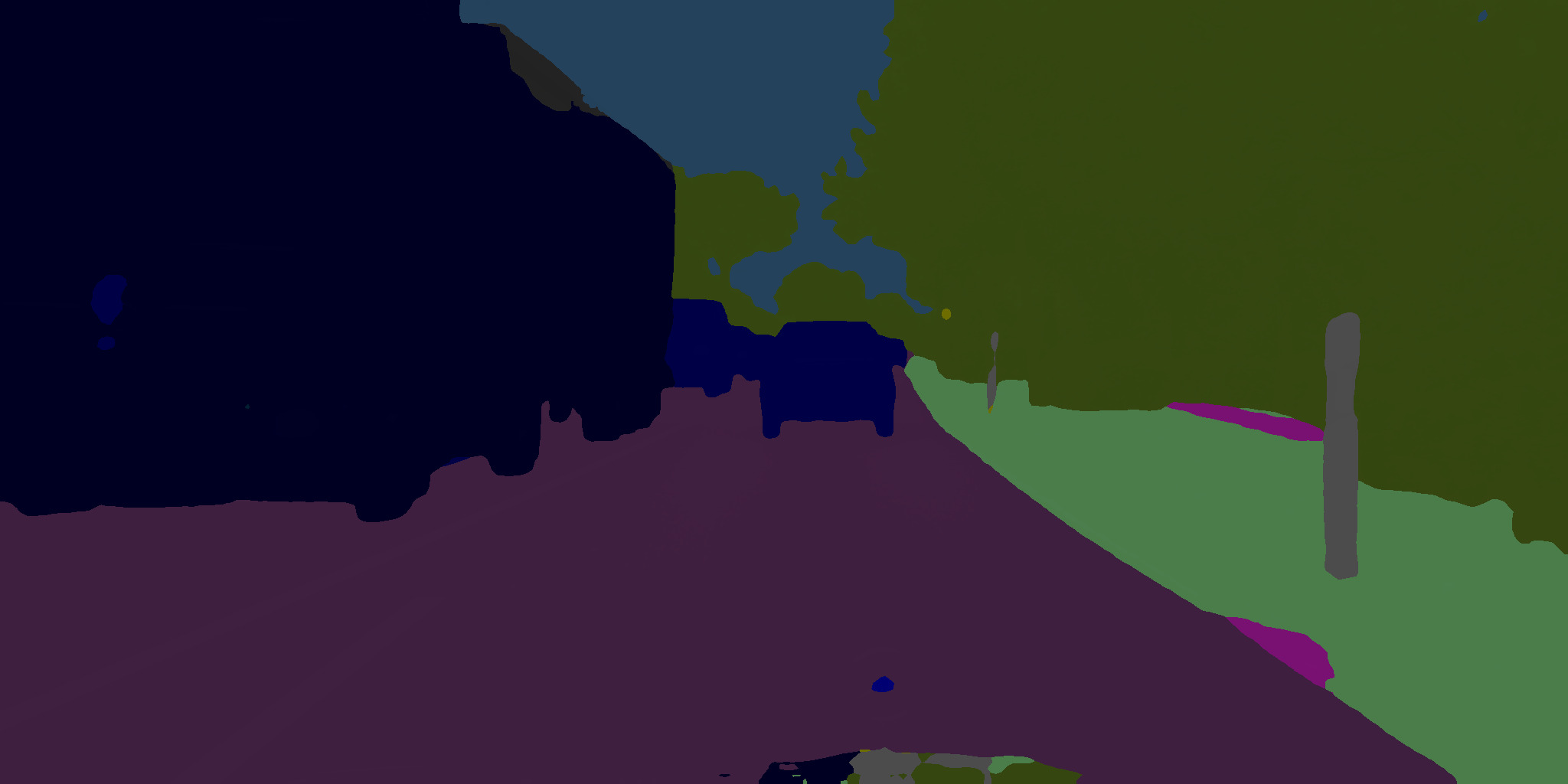}%
\end{subfigure}\hfill%
\begin{subfigure}{.4\columnwidth}
\includegraphics[height=2.5cm,width=\columnwidth]{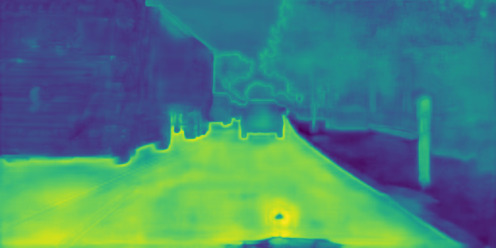}%
\end{subfigure}\hfill%
\begin{subfigure}{.4\columnwidth}
\includegraphics[height=2.5cm,width=\columnwidth]{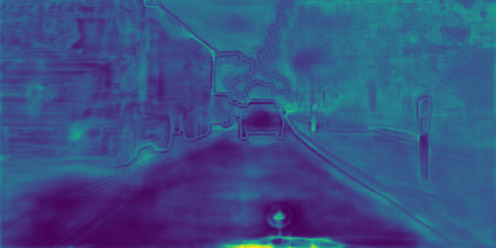}%
\end{subfigure}\hfill%
\begin{subfigure}{.4\columnwidth}
\includegraphics[height=2.5cm,width=\columnwidth]{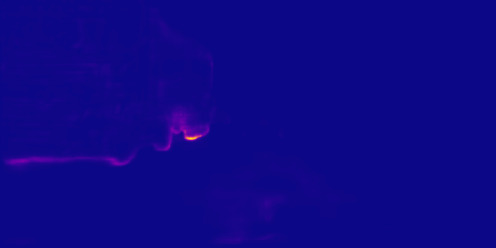}%
\end{subfigure}%
\\
\begin{subfigure}{.4\columnwidth}
\includegraphics[width=\columnwidth]{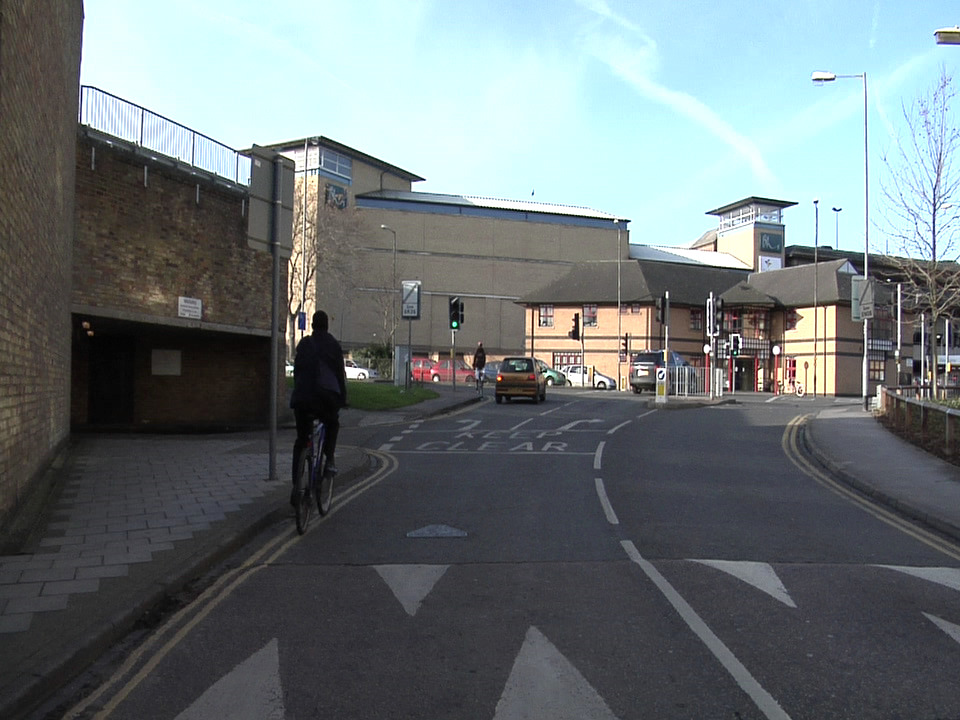}%
\caption*{input}
\end{subfigure}\hfill%
\begin{subfigure}{.4\columnwidth}
\includegraphics[width=\columnwidth]{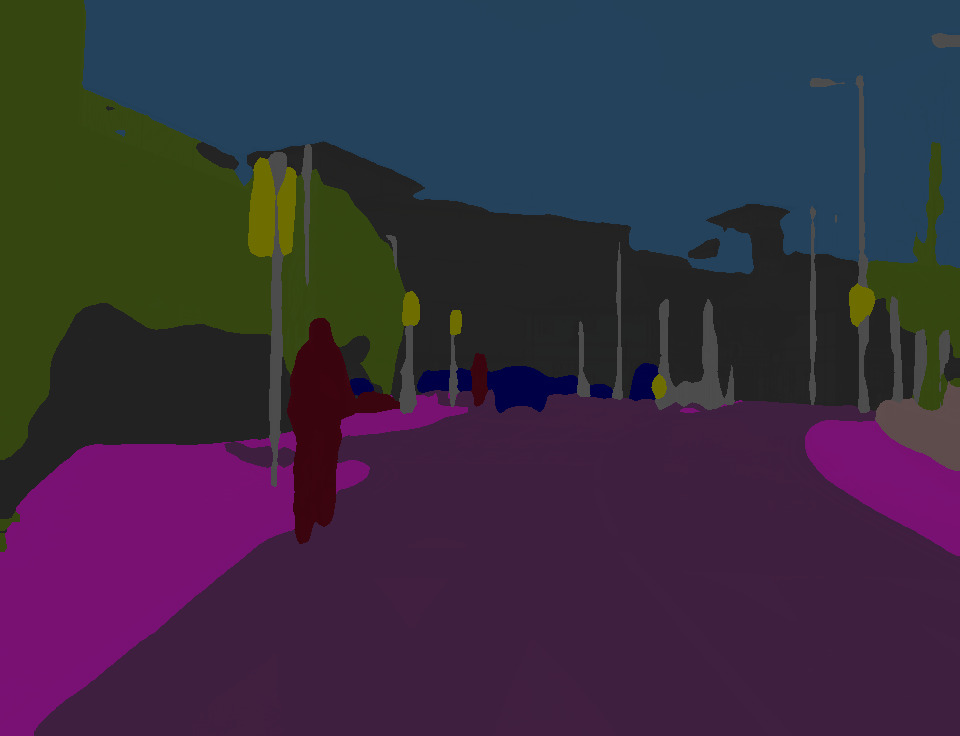}%
\caption*{predicted segmentation}
\end{subfigure}\hfill%
\begin{subfigure}{.4\columnwidth}
\includegraphics[width=\columnwidth]{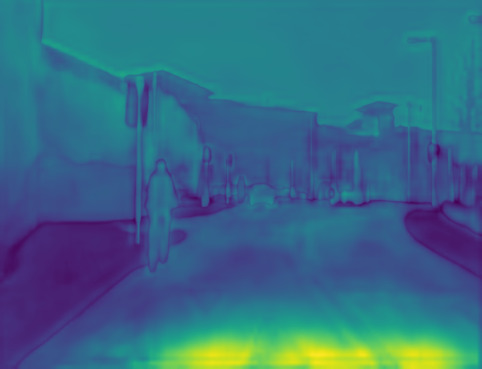}%
\caption*{epistemic uncertainty}
\end{subfigure}\hfill%
\begin{subfigure}{.4\columnwidth}
\includegraphics[width=\columnwidth]{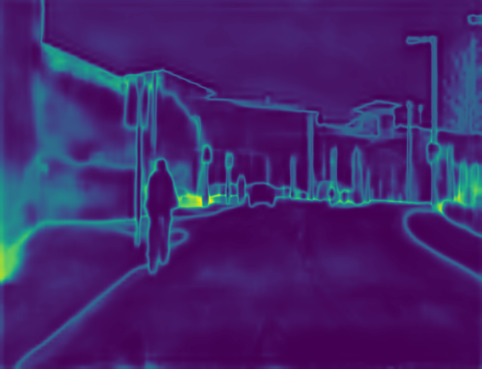}%
\caption*{aleatoric uncertainty}
\end{subfigure}\hfill%
\begin{subfigure}{.4\columnwidth}
\includegraphics[width=\columnwidth]{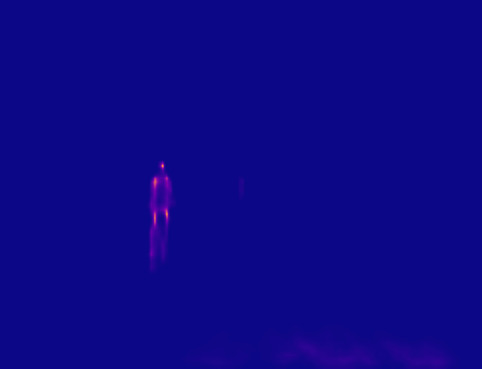}%
\caption*{class uncertainty}
\end{subfigure}%
\\
\caption{Examples of predictions from the proposed model for the CityScapes
  Dataset utilising the BiSeNetv1 backbone (top row) and the CamVid datasets
  using the PIDNet backbone (bottom row). From left to right, the columns
  represent the input, classification, epistemic uncertainty, aleatoric
  uncertainty, and class conditional epistemic uncertainty. For epistemic and
  aleatoric uncertainty, green regions indicate larger uncertainty, whilst red
  indicates greater uncertainty for class conditional uncertainty. For class
  conditional uncertainty, we show the variance for the classes ``truck'' and ``person''. Best viewed on computer screen.}
\label{fig:pred_uncertainty}
\end{figure*}

\par
For our inference procedure, we perform training on the existing models to obtain
our diagonal SWAG samples with the Ohem cross-entropy loss
\cite{shrivastava2016training}. We perform a total of 5,000 SGD iterations,
with a warmup of 1,000 iterations using a linear increase of the learning rate.
After this warmup period, the learning rate remains constant. The parameters are observed every 50 SGD iterations for the remainder of training for inference utilising diagonal SWAG. These recorded
parameters are then used to estimate the empirical covariance matrix for the final probabilistic convolutional layer. Our prior over our latent variables is a Gaussian distribution $\N(0, 1 \times 10^{-4})$, which is implemented using weight decay.
\par
After training is completed, the models are then compiled to the TensorRT format
to be used on the Jetson board for calculation of inference speed metrics.
During compilation, the models are converted to half-precision floating point
values to accelerate computation. To measure predictive speed, we time
the computation required for 1,000 forward passes. Our primary measures for
evaluating model predictive performance is the mean intersection over union
(mIOU) and the Macro and Micro F1 scores. Predictive speed is measured in
frames-per-second (fps) on the Jetson device, and compared against
traditional point estimate networks.
\subsection{Urban-Scene Datasets for Autonomous Driving}
One area where real-time predictions is required is autonomous vehicles.
Given the rapid development and need for uncertainty measures in this safety
critical scenario, we perform experimentation on the
 CamVid \cite{brostow2009semantic} and CityScapes \cite{cordts2016cityscapes}
 datasets where pre-trained base networks are available to form our feature extractor $f(:, \theta)$. For the
CamVid datasets, we perform experimentation adding the proposed \Gls{ep} uncertainty mechanism to ENet \cite{paszke2016enet} and PIDNet
\cite{xu2022pidnet}. For ENet models, UnMaxpooling operations in the decoder
were replaced with interpolation stages followed by a small finetuning stage to
accomodate the change. This change is required as TensorRT does not support the
unpooling operations. For CityScapes, a larger number of pre-trained models are
available, allowing us to perform experimentation using the previous models and
BiSeNetV1\cite{yu-2018-bisen}, BiSeNetv2 \cite{yu-2020-bisen-v2} and PPLiteSegT
and PPLiteSegB \cite{peng2022pp}. We summarise the predictive performance of
these models for the CityScapes and CamVid datasets in Tables
\ref{tab:performance_city} and \ref{tab:performance_cam} respectively, where
they are also compared against equivalent point estimate networks.

\par
From the results in~\Cref{tab:performance_city}, we see comparable performance
amongst the probabilistic models and their point estimate equivalents. Whilst
the predictive speed does decrease across the Bayesian models, we see that
predictive performance remains suitable for real time application.
Interestingly, we note that the models that have shown faster predictions on
traditional hardware such as ENet and BiSeNetV2 suffer from slower than expected predictive times. We attribute
this decrease in predictive performance on the Jetson hardware to the depth-wise
convolutional operations used in the BiSeNetv2 model and the excessive
interpolation stages in ENet not being as thoroughly
optimised within the TensorRT framework. This highlights the importance of
designing bespoke architectures for edge devices.
\begin{table}[!ht]
  \caption{Summary of predictive performance of real-time semantic segmentation
    methods on Jetson Xavier embedded hardware for CityScapes dataset.
    Uncertainty enabled models are given the "``Bayes-'' prefix.}
  \label{tab:performance_city}
  \begin{footnotesize}
    \begin{tabular}{l l l l l}
      Model & Macro-F1 & Micro-F1 & mIOU & fps
      \\
      \midrule\midrule[.1em]
      BiSeNetV1 & 0.8423& 0.9556& 0.7426 & 68.81\\
      Bayes-BiSeNetV1 & 0.8415& 0.9550 & 0.7408 & 57.359\\
      BiSeNetV2 & 0.8473& 0.9565 & 0.7476 & 49.502\\
      Bayes-BiSeNetV2 & 0.8456 & 0.9558 & 0.7451 & 57.359\\
      ENet & 0.7125  & 0.8976  & 0.5795  &  45.1638 \\
      Bayes-ENet  & 0.7125  & 0.8974  & 0.5795  & 30.709\\
      PIDNet & 0.9150  & 0.9705  & 0.8486 & 84.298 \\ %
      Bayes-PIDNet & 0.9129  & 0.9695  & 0.8454 & 69.367 \\ %
      PPLiteSegT & 0.8534  & 0.9558  & 0.7578 & 72.5234 \\ %
      Bayes-PPLiteSegT & 0.8481 & 0.9536  & 0.7501 &  65.931 \\ %
      PPLiteSegB & 0.8666   & 0.9597  & 0.7750& 56.9437\\ %
      Bayes-PPLiteSegB & 0.8603  & 0.9571  & 0.7656& 52.022  \\ %
      \midrule
    \end{tabular}
  \end{footnotesize}
\end{table}
\begin{table}[!ht]
  \caption{Summary of predictive performance of real-time semantic segmentation
    methods on Jetson Xavier embedded hardware for CamVid dataset.
    Uncertainty enabled models are given the "``Bayes-'' prefix.}
  \label{tab:performance_cam}
  \begin{footnotesize}
    \begin{tabular}{l l l l l}
      Model &  Macro-F1 & Micro-F1&  mIOU & fps
      \\
      \midrule\midrule[.1em]
      ENet & 0.6932 & 0.9116& 0.5950 & 42.2916 \\
      Bayes-ENet & 0.6702& 0.8990 & 0.5698 & 30.825\\
      PIDNet &  0.8683 & 0.9485 &  0.7847 & 73.7831\\ %
      Bayes-PIDNet & 0.8661 & 0.9471 & 0.7816&  59.978\\ %
      \midrule
    \end{tabular}
  \end{footnotesize}
\end{table}
\par
\begin{figure}%
\centering
\begin{subfigure}[t]{.45\columnwidth}
\includegraphics[height=2.3cm,width=\columnwidth]{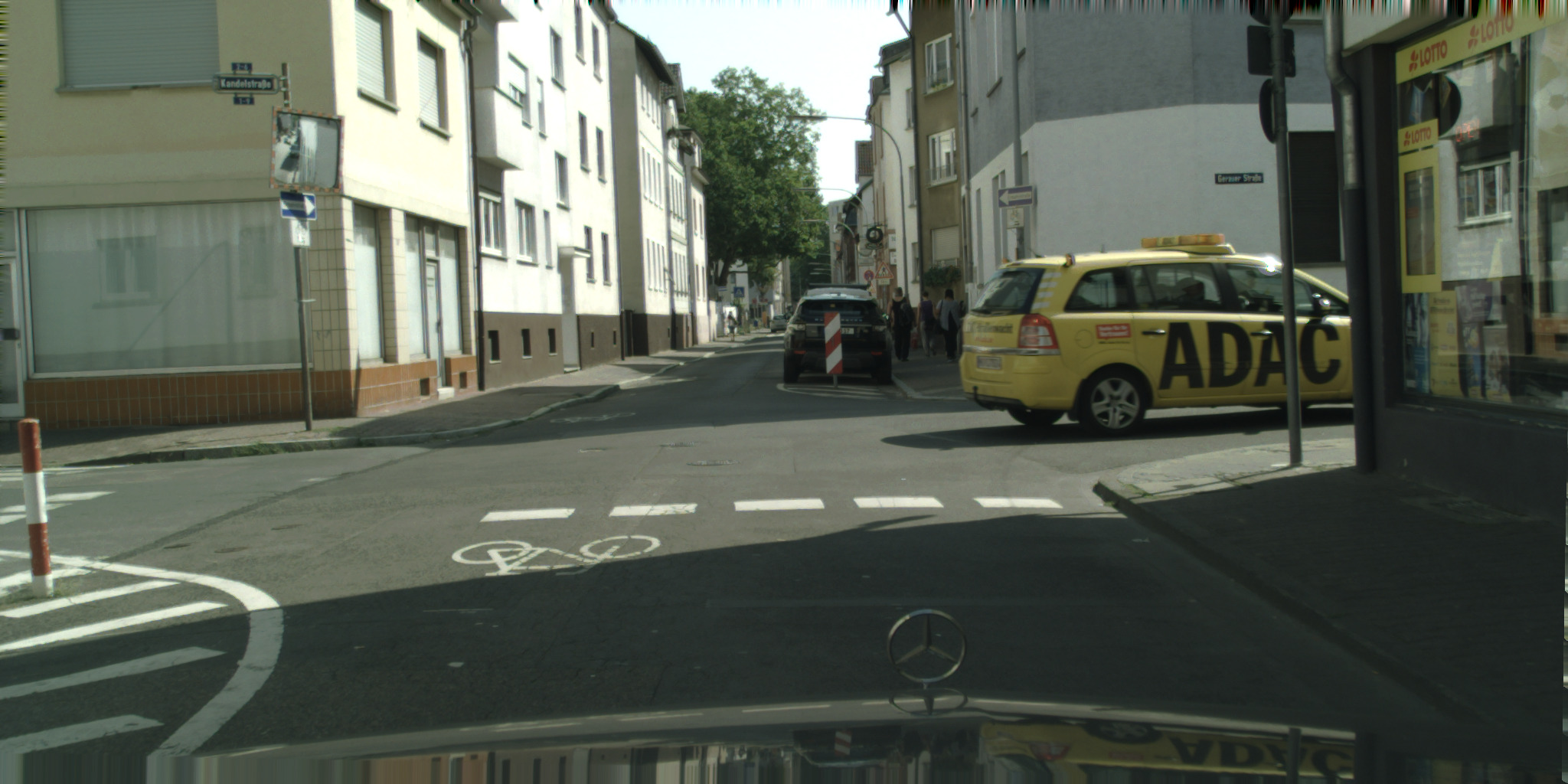}%
\caption{Input}
\end{subfigure}\hspace*{0.1cm}%
\begin{subfigure}[t]{.45\columnwidth}
\includegraphics[height=2.3cm,width=\columnwidth]{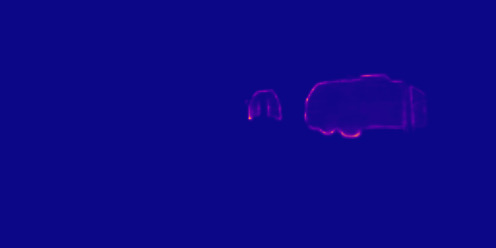}%
\caption{``car'' std.}
\end{subfigure}%
\\
\begin{subfigure}[t]{.45\columnwidth}
\includegraphics[height=2.3cm,width=\columnwidth]{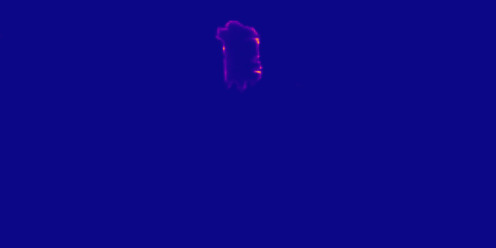}%
\caption{``vegetation'' std.}
\end{subfigure}\hspace*{0.1cm}%
\begin{subfigure}[t]{.45\columnwidth}
\includegraphics[height=2.3cm,width=\columnwidth]{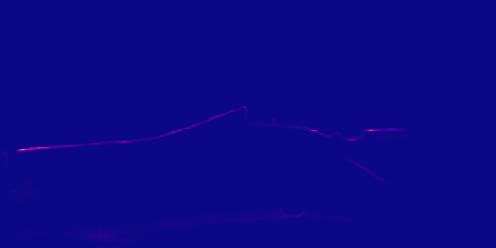}%
\caption{``road'' std.}
\end{subfigure}%
\caption{Visualisation of class-conditional uncertainty for a sample from
  CityScapes dataset. Red regions indicate an increase in uncertainty.
  Uncertainties presented as variance. Predictions performed with PIDNet.}
\label{fig:pred_city_class_uncertainty}
\end{figure}

With our predictive performance measured quantitatively, we now investigate the
uncertainty measures qualitatively. In~\Cref{rec:comp_pred_uncertainty}, we
state how the proposed modelling method is capable of producing measures for
aleatoric, epistemic and class conditional epistemic
uncertainty.~\Cref{fig:pred_uncertainty} illustrates the predictive performance
over CityScapes and CamVid datasets and the relevant uncertainty
measures. We can see from these figures that the measured epistemic uncertainty
from entropy in Gaussian representation of our output (Eqn \ref{eq:entropy_gaussian}) is largest within the objects of interest, whilst the aleatoric uncertainty
 measured from the entropy of the final categorical distribution  (Eqn \ref{eq:aleatoric_uncertainty}) is concentrated
 around the edges of these objects. We further see from the class
conditional uncertainty measures that when targeting individual classes, the
uncertainty is targeted towards the edge of the objects for the class of
interest. We explore additional examples of the class-conditional uncertainty
measures in Figure \ref{fig:pred_city_class_uncertainty} to further verify this result.

\subsection{Semantic Segmentation of General Objects}
\begin{figure*}%
\centering
\begin{subfigure}{.4\columnwidth}
\includegraphics[width=\columnwidth]{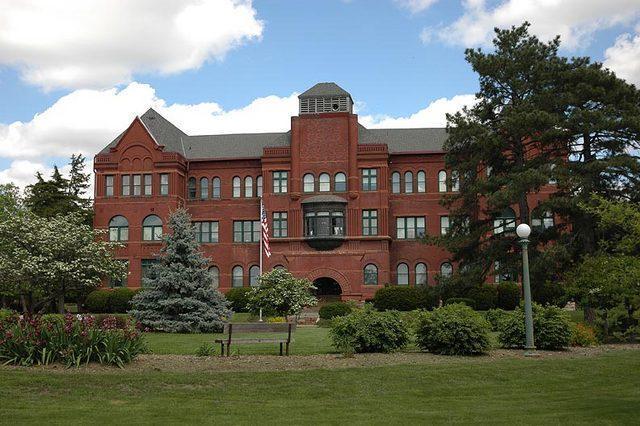}%
\end{subfigure}\hfill%
\begin{subfigure}{.4\columnwidth}
\includegraphics[width=\columnwidth]{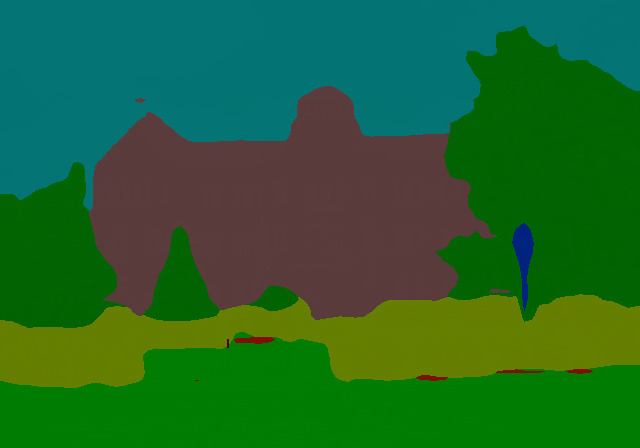}%
\end{subfigure}\hfill%
\begin{subfigure}{.4\columnwidth}
\includegraphics[width=\columnwidth]{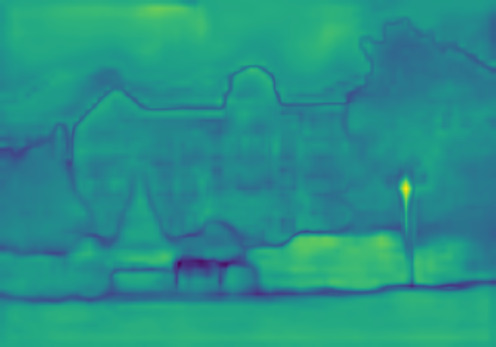}%
\end{subfigure}\hfill%
\begin{subfigure}{.4\columnwidth}
\includegraphics[width=\columnwidth]{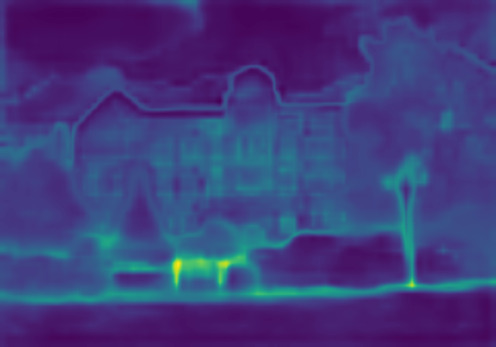}%
\end{subfigure}\hfill%
\begin{subfigure}{.4\columnwidth}
\includegraphics[width=\columnwidth]{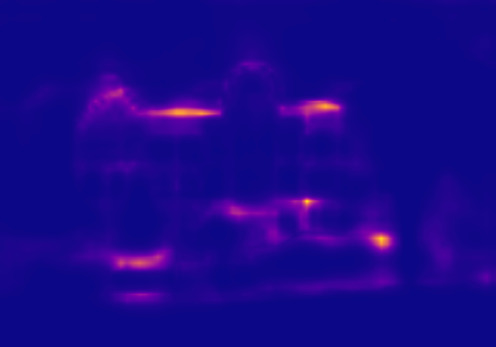}%
\end{subfigure}%
\\
\begin{subfigure}{.4\columnwidth}
\includegraphics[width=\columnwidth]{figs/bisenetv1_figs/CocoStuff/decision/im.jpg}%
\caption*{input}
\end{subfigure}\hfill%
\begin{subfigure}{.4\columnwidth}
\includegraphics[width=\columnwidth]{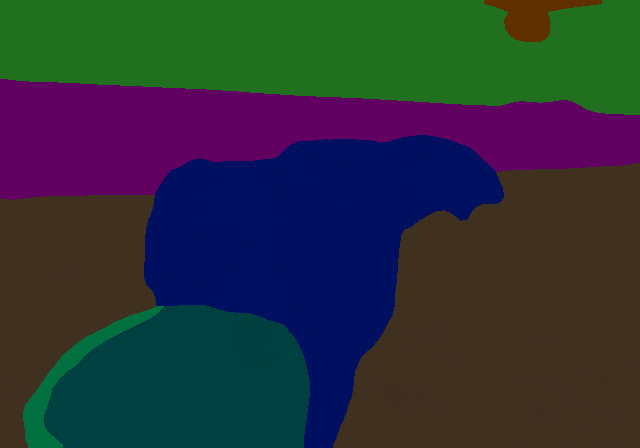}%
\caption*{predicted segmentation}
\end{subfigure}\hfill%
\begin{subfigure}{.4\columnwidth}
\includegraphics[width=\columnwidth]{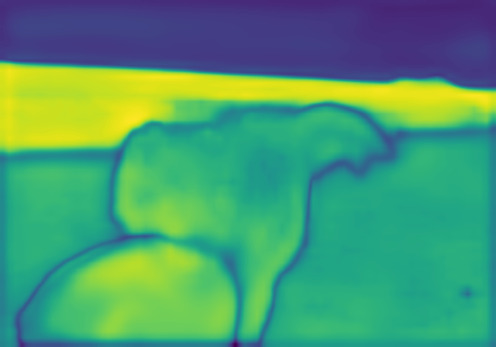}%
\caption*{epistemic uncertainty}
\end{subfigure}\hfill%
\begin{subfigure}{.4\columnwidth}
\includegraphics[width=\columnwidth]{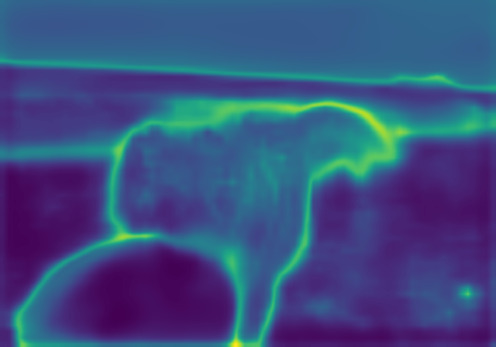}%
\caption*{aleatoric uncertainty}
\end{subfigure}\hfill%
\begin{subfigure}{.4\columnwidth}
\includegraphics[width=\columnwidth]{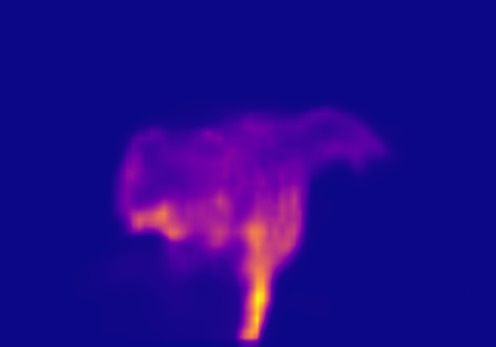}%
\caption*{class uncertainty}
\end{subfigure}%
\caption{Examples of predictions from the proposed model for the ADE20k Dataset
  (top row) and the CoCoStuff dataset (bottom row) when utilising the BiSeNetv1
  backbone. From left to right, the columns represent the input, classification,
  epistemic uncertainty, aleatoric uncertainty, and class conditional epistemic
  uncertainty. For epistemic and aleatoric uncertainty, green regions indicate
  larger uncertainty, whilst red indicates greater uncertainty for class
  conditional uncertainty. For class conditional uncertainty, we show the
  variance for the classes ``building'' and ``dog''. Best viewed on computer screen.}
\label{fig:pred_object_uncertainty}
\end{figure*}

\begin{figure}%
\centering
\begin{subfigure}[t]{.4\columnwidth}
\includegraphics[width=\columnwidth]{figs/class_bisenetv2/im.jpg}%
\caption{Input}
\end{subfigure}\hspace*{0.1cm}%
\begin{subfigure}[t]{.4\columnwidth}
\includegraphics[width=\columnwidth]{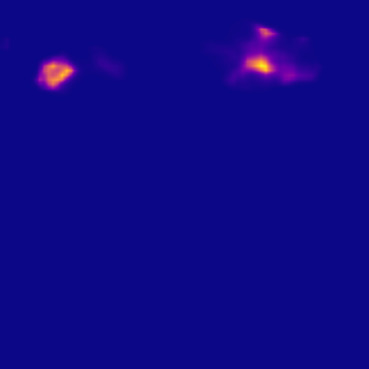}%
\caption{``car'' std.}
\end{subfigure}%
\\
\begin{subfigure}[t]{.4\columnwidth}
\includegraphics[width=\columnwidth]{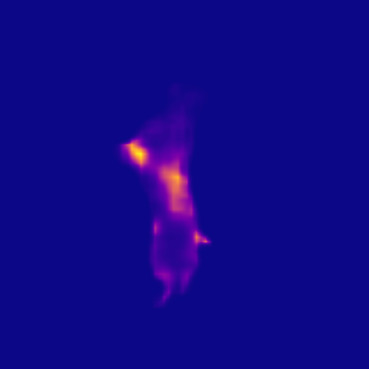}%
\caption{``dog'' std.}
\end{subfigure}\hspace*{0.1cm}%
\begin{subfigure}[t]{.4\columnwidth}
\includegraphics[width=\columnwidth]{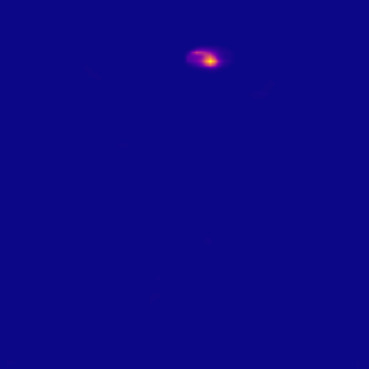}%
\caption{``frisbee'' std.}
\end{subfigure}%
\caption{Visualisation of class-conditional uncertainty for a sample from
  CoCoStuff dataset. Red regions indicate an increase in uncertainty.
  Uncertainty presented as variance. Predictions performed with BiSeNetV2.}
\label{fig:pred_coco_class_uncertainty}
\end{figure}

Robotics and human computer interaction systems are another class of practical
application where semantic segmentation methods are required to operate in
real-time on edge devices. Given the complexity of operating with humans or the
dynamic environments robotics systems encounter, it is important that segmentation methods used within
these scenarios are able to reason about uncertainty present. We perform
additional experimentation using the challenging ADE20k \cite{zhou2017scene} and
CoCoStuff \cite{caesar2018coco} datasets to assert the suitability of the
proposed probabilistic module for data seen within these scenarios. These
experiments build upon the available pre-trained BiSeNetV1 and BiSeNetV2 models to form $f(:, \theta)$,
with results shown in Tables \ref{tab:pred_coco} and \ref{tab:pred_ade}.
Visualisations of uncertainty measures is shown in
Figure \ref{fig:pred_object_uncertainty}, and examples of class conditional uncertainty
in Figure \ref{fig:pred_coco_class_uncertainty}.

\begin{table}[!ht]
  \caption{Summary of predictive performance of real-time semantic segmentation
    methods on Jetson Xavier embedded hardware for CoCoStuff dataset.}
  \label{tab:pred_coco}
  \begin{footnotesize}
    \begin{tabular}{l l l l l}
      Model &  Macro-F1 & Micro-F1&  mIOU & fps
      \\
      \midrule\midrule[.1em]
      BiSeNetV1 & 0.4254 & 0.6295 & 0.3077 & 72.4150 \\
      Bayes-BiSeNetV1 & 0.4256  & 0.6300 & 0.3079& 43.3340 \\
      BiSeNetV2 & 0.4042 & 0.6025& 0.2851 & 58.6345 \\
      Bayes-BiSeNetV2 & 0.4047 & 0.6027& 0.2855 & 41.2032 \\
      \midrule
    \end{tabular}
  \end{footnotesize}
\end{table}
\begin{table}[!ht]
  \caption{Summary of predictive performance of real-time semantic segmentation
    methods on Jetson Xavier embedded hardware for ADE20k dataset.}
  \label{tab:pred_ade}
  \begin{footnotesize}
    \begin{tabular}{l l l l l}
      Model &  Macro-F1 & Micro-F1&  mIOU & fps
      \\
      \midrule\midrule[.1em]
      BiSeNetV1 & 0.4859 & 0.7643 & 0.3513 & 62.4800 \\
      Bayes-BiSeNetV1 & 0.4861 & 0.7642& 0.3514& 54.2450 \\
      BiSeNetV2 & 0.4490 & 0.7593 & 0.3211 & 59.5024 \\
      Bayes-BiSeNetV2 & 0.4494 & 0.7591& 0.3215 & 41.8055 \\
      \midrule
    \end{tabular}
  \end{footnotesize}
\end{table}
These results further demonstrate the ability of the proposed uncertainty
mechanism to provide comparable predictive accuracy to the corresponding point-estimate networks on
embedded hardware, whilst also providing a range of meaningful uncertainty
estimates. We see in Figure \ref{fig:pred_coco_class_uncertainty} that class
conditional accuracy has started to become much more diffuse, though is still
concentrated around the edges of the objects of interest. This can be attributed
to the greater complexity and diversity offered within the ADE20k and CoCoStuff
datasets, and highlights the importance of these uncertainty measures for these
challenging scenarios.

\section{Discussion and Future Research}
\label{sec:discussion}
Experimental results in Section \ref{sec:experiments} have demonstrated how
the proposed uncertainty propegation method can provide meaningful uncertainty
estimates when included with a variety of existing real-time segmentation
networks. From these results we highlight areas for future research and how to
further enhance the ability to capture uncertainty within real-time
semantic segmentation models.
\par
The proposed approach builds upon existing segmentation models to enable them to
capture uncertainty present within parameters in the classification layer and
propegate these uncertainties to meaningful predictive measurements. Since
the incorporation of the proposed uncertainty mechanism is included
post-training of the feature extractor $f(;, \theta)$, any uncertainty present during training of
the base network is not explicitly captured. Recent methods have proposed
training mechanisms to fit semantic-segmentation models
that incorporate aleatoric uncertainty into the training procedure
\cite{fleuret2021uncertainty,wang2021uncertainty,zhou2022uncertainty}.
Incorporation of these training schemes with the proposed uncertainty
propegation module could allow meaningful epistemic uncertainty estimates to be found whilst encouraging
the base network to have a more accurate representation of aleatoric uncertainty
present for resource constrained hardware.
\par
Whilst the vast majority of work on modelling uncertainty within deep-learning
has focused on finding predictive uncertainty estimates, there has been
relatively little research on how best to utilise these measures in practice
despite the existence of a rich and mature theory to address these important
challenges\cite{degroot2005optimal,kochenderfer2022algorithms}. Given the rising
prevalence of semantic segmentaion systems to safety critical environments such
as autonomous vehicles, medical imaging or human-computer interaction, it is
crucial that sensible and safe decisions are made by the machine learning
system. Optimal and safe decisions require the incorporation of
meaningful uncertainty information. The sensitivity to context of different
application poses further challenges associated with optimal and
safe decision making for automated systems. Given the need to process and
interpret uncertainty information present, the additional processing of this
information will incur a computational overhead. This computational burden is
especially important for real-time embedded systems, and we highlight as an
important area for future research in practical uncertainty aware semantic segmentation systems.

\section{Conclusion}
\label{sec:conclusion}
Within this research, we targeted the need for meaningful uncertainty information
for semantic segmentation on resource constrained hardware. We proposed
a combination of a deterministic feature extractor with a probabilistic regression
module, that can be combined with an \Gls{ep} softmax module to generate final
predictive outputs and uncertainties. We demonstrate how the proposed probabilistic module can be included with existing pre-trained networks.
Evaluation of these models on embedded hardware using multiple datasets demonstrate how the proposed probabilistic scheme can mitigate the
expensive compute traditionally required for a complete probabilistic treatment of deep segmentation
models to provide real-time performance on resource constrained hardware. Qualitative evaluation of obtained uncertainty
measures demonstrate how the different forms of uncertainty obtained can be used to identify different regions of interest.

\bibliographystyle{IEEEbib}
\bibliography{ref}

\begin{thebibliography}{10}

\bibitem{brostow2009semantic}
Gabriel~J Brostow, Julien Fauqueur, and Roberto Cipolla,
\newblock ``Semantic object classes in video: A high-definition ground truth
  database,''
\newblock {\em Pattern Recognition Letters}, vol. 30, no. 2, pp. 88--97, 2009.

\bibitem{caesar2018coco}
Holger Caesar, Jasper Uijlings, and Vittorio Ferrari,
\newblock ``Coco-stuff: Thing and stuff classes in context,''
\newblock in {\em Proceedings of the IEEE Conference on Computer Vision and
  Pattern Recognition (CVPR)}, June 2018.

\bibitem{cordts2016cityscapes}
Marius Cordts, Mohamed Omran, Sebastian Ramos, Timo Rehfeld, Markus Enzweiler,
  Rodrigo Benenson, Uwe Franke, Stefan Roth, and Bernt Schiele,
\newblock ``The cityscapes dataset for semantic urban scene understanding,''
\newblock in {\em Proceedings of the IEEE conference on computer vision and
  pattern recognition}, 2016, pp. 3213--3223.

\bibitem{milioto2018real}
Andres Milioto, Philipp Lottes, and Cyrill Stachniss,
\newblock ``Real-time semantic segmentation of crop and weed for precision
  agriculture robots leveraging background knowledge in cnns,''
\newblock in {\em 2018 IEEE international conference on robotics and automation
  (ICRA)}. IEEE, 2018, pp. 2229--2235.

\bibitem{zhou2017scene}
Bolei Zhou, Hang Zhao, Xavier Puig, Sanja Fidler, Adela Barriuso, and Antonio
  Torralba,
\newblock ``Scene parsing through ade20k dataset,''
\newblock in {\em Proceedings of the IEEE Conference on Computer Vision and
  Pattern Recognition}, 2017.

\bibitem{mukhoti2018evaluating}
Jishnu Mukhoti and Yarin Gal,
\newblock ``Evaluating bayesian deep learning methods for semantic
  segmentation,''
\newblock {\em arXiv preprint arXiv:1811.12709}, 2018.

\bibitem{dechesne2021bayesian}
Clément Dechesne, Pierre Lassalle, and Sébastien Lefèvre,
\newblock ``Bayesian u-net: Estimating uncertainty in semantic segmentation of
  earth observation images,''
\newblock {\em Remote Sensing}, vol. 13, no. 19, 2021.

\bibitem{kampffmeyer2016semantic}
Michael Kampffmeyer, Arnt-Borre Salberg, and Robert Jenssen,
\newblock ``Semantic segmentation of small objects and modeling of uncertainty
  in urban remote sensing images using deep convolutional neural networks,''
\newblock in {\em Proceedings of the IEEE conference on computer vision and
  pattern recognition workshops}, 2016, pp. 1--9.

\bibitem{sivaprasad2021uncertainty}
Prabhu~Teja Sivaprasad and Francois Fleuret,
\newblock ``Uncertainty reduction for model adaptation in semantic
  segmentation,''
\newblock in {\em 2021 Ieee/Cvf Conference On Computer Vision And Pattern
  Recognition, Cvpr 2021}. IEEE, 2021, number CONF, pp. 9608--9618.

\bibitem{zheng2021rectifying}
Zhedong Zheng and Yi~Yang,
\newblock ``Rectifying pseudo label learning via uncertainty estimation for
  domain adaptive semantic segmentation,''
\newblock {\em International Journal of Computer Vision}, vol. 129, no. 4, pp.
  1106--1120, 2021.

\bibitem{wilson2016deep}
Andrew~Gordon Wilson, Zhiting Hu, Ruslan Salakhutdinov, and Eric~P Xing,
\newblock ``Deep kernel learning,''
\newblock in {\em Artificial intelligence and statistics}. PMLR, 2016, pp.
  370--378.

\bibitem{snoek2015scalable}
Jasper Snoek, Oren Rippel, Kevin Swersky, Ryan Kiros, Nadathur Satish,
  Narayanan Sundaram, Mostofa Patwary, Mr~Prabhat, and Ryan Adams,
\newblock ``Scalable bayesian optimization using deep neural networks,''
\newblock in {\em International conference on machine learning}. PMLR, 2015,
  pp. 2171--2180.

\bibitem{chen2022vision}
Zhe Chen, Yuchen Duan, Wenhai Wang, Junjun He, Tong Lu, Jifeng Dai, and
  Yu~Qiao,
\newblock ``Vision transformer adapter for dense predictions,''
\newblock {\em arXiv preprint arXiv:2205.08534}, 2022.

\bibitem{yuan2019segmentation}
Yuhui Yuan, Xiaokang Chen, Xilin Chen, and Jingdong Wang,
\newblock ``Segmentation transformer: Object-contextual representations for
  semantic segmentation,''
\newblock 2020.

\bibitem{wang2022internimage}
Wenhai Wang, Jifeng Dai, Zhe Chen, Zhenhang Huang, Zhiqi Li, Xizhou Zhu,
  Xiaowei Hu, Tong Lu, Lewei Lu, Hongsheng Li, et~al.,
\newblock ``Internimage: Exploring large-scale vision foundation models with
  deformable convolutions,''
\newblock {\em arXiv preprint arXiv:2211.05778}, 2022.

\bibitem{fu2020scene}
Jun Fu, Jing Liu, Jie Jiang, Yong Li, Yongjun Bao, and Hanqing Lu,
\newblock ``Scene segmentation with dual relation-aware attention network,''
\newblock {\em IEEE Transactions on Neural Networks and Learning Systems}, vol.
  32, no. 6, pp. 2547--2560, 2020.

\bibitem{long2015fully}
Jonathan Long, Evan Shelhamer, and Trevor Darrell,
\newblock ``Fully convolutional networks for semantic segmentation,''
\newblock in {\em Proceedings of the IEEE conference on computer vision and
  pattern recognition}, 2015, pp. 3431--3440.

\bibitem{paszke2016enet}
Adam Paszke, Abhishek Chaurasia, Sangpil Kim, and Eugenio Culurciello,
\newblock ``Enet: A deep neural network architecture for real-time semantic
  segmentation,''
\newblock {\em arXiv preprint arXiv:1606.02147}, 2016.

\bibitem{he2016deep}
Kaiming He, Xiangyu Zhang, Shaoqing Ren, and Jian Sun,
\newblock ``Deep residual learning for image recognition,''
\newblock in {\em Proceedings of the IEEE conference on computer vision and
  pattern recognition}, 2016, pp. 770--778.

\bibitem{wang2021swiftnet}
Haochen Wang, Xiaolong Jiang, Haibing Ren, Yao Hu, and Song Bai,
\newblock ``Swiftnet: Real-time video object segmentation,''
\newblock in {\em Proceedings of the IEEE/CVF Conference on Computer Vision and
  Pattern Recognition}, 2021, pp. 1296--1305.

\bibitem{yu-2018-bisen}
Changqian Yu, Jingbo Wang, Chao Peng, Changxin Gao, Gang Yu, and Nong Sang,
\newblock ``Bisenet: Bilateral segmentation network for real-time semantic
  segmentation,''
\newblock {\em CoRR}, 2018.

\bibitem{yu-2020-bisen-v2}
Changqian Yu, Changxin Gao, Jingbo Wang, Gang Yu, Chunhua Shen, and Nong Sang,
\newblock ``Bisenet v2: Bilateral network with guided aggregation for real-time
  semantic segmentation,''
\newblock {\em CoRR}, 2020.

\bibitem{xu2022pidnet}
Jiacong Xu, Zixiang Xiong, and Shankar~P Bhattacharyya,
\newblock ``Pidnet: A real-time semantic segmentation network inspired from pid
  controller,''
\newblock {\em arXiv preprint arXiv:2206.02066}, 2022.

\bibitem{kendall2017bayes}
Alex Kendall, Vijay Badrinarayanan, and Roberto Cipolla,
\newblock ``Bayesian segnet: Model uncertainty in deep convolutional
  encoder-decoder architectures for scene understanding,''
\newblock in {\em British Machine Vision Conference 2017, {BMVC} 2017, London,
  UK, September 4-7, 2017}. 2017, {BMVA} Press.

\bibitem{kendall2017uncertainties}
Alex Kendall and Yarin Gal,
\newblock ``What uncertainties do we need in bayesian deep learning for
  computer vision?,''
\newblock {\em Advances in neural information processing systems}, vol. 30,
  2017.

\bibitem{van2020uncertainty}
Joost Van~Amersfoort, Lewis Smith, Yee~Whye Teh, and Yarin Gal,
\newblock ``Uncertainty estimation using a single deep deterministic neural
  network,''
\newblock in {\em International conference on machine learning}. PMLR, 2020,
  pp. 9690--9700.

\bibitem{mukhoti2021deep}
Jishnu Mukhoti, Joost van Amersfoort, Philip~HS Torr, and Yarin Gal,
\newblock ``Deep deterministic uncertainty for semantic segmentation,''
\newblock {\em arXiv preprint arXiv:2111.00079}, 2021.

\bibitem{huang2018efficient}
Po-Yu Huang, Wan-Ting Hsu, Chun-Yueh Chiu, Ting-Fan Wu, and Min Sun,
\newblock ``Efficient uncertainty estimation for semantic segmentation in
  videos,''
\newblock in {\em Proceedings of the European Conference on Computer Vision
  (ECCV)}, 2018, pp. 520--535.

\bibitem{qu2021bayesian}
Chao Qu, Wenxin Liu, and Camillo~J Taylor,
\newblock ``Bayesian deep basis fitting for depth completion with
  uncertainty,''
\newblock in {\em Proceedings of the IEEE/CVF international conference on
  computer vision}, 2021, pp. 16147--16157.

\bibitem{frey1999variational}
Brendan~J Frey and Geoffrey~E Hinton,
\newblock ``Variational learning in nonlinear gaussian belief networks,''
\newblock {\em Neural Computation}, vol. 11, no. 1, pp. 193--213, 1999.

\bibitem{gast2018light}
Jochen Gast and Stefan Roth,
\newblock ``Lightweight probabilistic deep networks,''
\newblock {\em CoRR}, 2018.

\bibitem{murphey2012machine}
kevin Murphey,
\newblock {\em Machine learning, a probabilistic perspective},
\newblock MIT Press, Cambridge, MA, 2012.

\bibitem{maddox2019simple}
Wesley~J. Maddox, Pavel Izmailov, Timur Garipov, Dmitry~P. Vetrov, and
  Andrew~Gordon Wilson,
\newblock ``A simple baseline for bayesian uncertainty in deep learning,''
\newblock in {\em Advances in Neural Information Processing Systems 32: Annual
  Conference on Neural Information Processing Systems 2019, NeurIPS 2019,
  December 8-14, 2019, Vancouver, BC, Canada}, Hanna~M. Wallach, Hugo
  Larochelle, Alina Beygelzimer, Florence d'Alch{\'{e}}{-}Buc, Emily~B. Fox,
  and Roman Garnett, Eds., 2019, pp. 13132--13143.

\bibitem{diaz2013existence}
Elo{\'\i}sa D{\'\i}az-Franc{\'e}s and Francisco~J Rubio,
\newblock ``On the existence of a normal approximation to the distribution of
  the ratio of two independent normal random variables,''
\newblock {\em Statistical Papers}, vol. 54, no. 2, pp. 309--323, 2013.

\bibitem{smith2018understanding}
Lewis Smith and Yarin Gal,
\newblock ``Understanding measures of uncertainty for adversarial example
  detection,''
\newblock in {\em Uncertainty in AI}, 2018.

\bibitem{shrivastava2016training}
Abhinav Shrivastava, Abhinav Gupta, and Ross Girshick,
\newblock ``Training region-based object detectors with online hard example
  mining,''
\newblock in {\em Proceedings of the IEEE conference on computer vision and
  pattern recognition}, 2016, pp. 761--769.

\bibitem{peng2022pp}
Juncai Peng, Yi~Liu, Shiyu Tang, Yuying Hao, Lutao Chu, Guowei Chen, Zewu Wu,
  Zeyu Chen, Zhiliang Yu, Yuning Du, et~al.,
\newblock ``Pp-liteseg: A superior real-time semantic segmentation model,''
\newblock {\em arXiv preprint arXiv:2204.02681}, 2022.

\bibitem{fleuret2021uncertainty}
Francois Fleuret et~al.,
\newblock ``Uncertainty reduction for model adaptation in semantic
  segmentation,''
\newblock in {\em Proceedings of the IEEE/CVF Conference on Computer Vision and
  Pattern Recognition}, 2021, pp. 9613--9623.

\bibitem{wang2021uncertainty}
Yuxi Wang, Junran Peng, and ZhaoXiang Zhang,
\newblock ``Uncertainty-aware pseudo label refinery for domain adaptive
  semantic segmentation,''
\newblock in {\em Proceedings of the IEEE/CVF International Conference on
  Computer Vision}, 2021, pp. 9092--9101.

\bibitem{zhou2022uncertainty}
Qianyu Zhou, Zhengyang Feng, Qiqi Gu, Guangliang Cheng, Xuequan Lu, Jianping
  Shi, and Lizhuang Ma,
\newblock ``Uncertainty-aware consistency regularization for cross-domain
  semantic segmentation,''
\newblock {\em Computer Vision and Image Understanding}, vol. 221, pp. 103448,
  2022.

\bibitem{degroot2005optimal}
Morris~H DeGroot,
\newblock {\em Optimal statistical decisions},
\newblock John Wiley \& Sons, 2005.

\bibitem{kochenderfer2022algorithms}
Mykel~J Kochenderfer, Tim~A Wheeler, and Kyle~H Wray,
\newblock {\em Algorithms for decision making},
\newblock MIT press, 2022.

\end{thebibliography}

\end{document}